*Article*

# Spatiotemporal Recurrent Convolutional Networks for Traffic Prediction in Transportation Networks

**Haiyang Yu** [1], **Zhihai Wu** [1], **Shuqin Wang** [2], **Yunpeng Wang** [1] **and Xiaolei Ma** [1,*]

[1] School of Transportation Science and Engineering, Beijing Key Laboratory for Cooperative Vehicle Infrastructure System and Safety Control, Beihang University, Beijing 100191, China; hyyu@buaa.edu.cn (H.Y.); zhihaiwu@buaa.edu.cn (Z.W.); ypwang@buaa.edu.cn (Y.W.); xiaolei@buaa.edu.cn (X.M.);
[2] Passenger Vehicle EE Development Department, China FAW R&D Center; wangshuqin@rdc.faw.com.cn (S.W.);
* Correspondence: xiaolei@buaa.edu.cn; Tel.: +86-13911966016

**Abstract:** Predicting large-scale transportation network traffic has become an important and challenging topic in recent decades. Inspired by the domain knowledge of motion prediction, in which the future motion of an object can be predicted based on previous scenes, we propose a network grid representation method that can retain the fine-scale structure of a transportation network. Network-wide traffic speeds are converted into a series of static images and input into a novel deep architecture, namely, spatiotemporal recurrent convolutional networks (SRCNs), for traffic forecasting. The proposed SRCNs inherit the advantages of deep convolutional neural networks (DCNNs) and long short-term memory (LSTM) neural networks. The spatial dependencies of network-wide traffic can be captured by DCNNs, and the temporal dynamics can be learned by LSTMs. An experiment on a Beijing transportation network with 278 links demonstrates that SRCNs outperform other deep learning-based algorithms in both short-term and long-term traffic prediction.

## 1. Introduction

Predicting large-scale network-wide traffic is a vital and challenging topic for transportation researchers. Traditional traffic prediction studies either relied on theoretical mathematical models to describe the traffic flow properties (i.e., model-driven approaches) or employed a variety of statistical learning and artificial intelligence algorithms (i.e., data-driven approaches). Model-driven approaches are criticized for strong assumptions and thus are inappropriate for application to real scenarios. Data-driven approaches have become increasingly popular due to the extensive deployment of traffic sensors and advanced data processing technologies. However, the majority of existing approaches tend to design and validate the proposed algorithms on expressways or several intersections [1, 2]. As discussed in [3], most traffic prediction methods that consider both spatial correlations and temporal correlations limit the input dimensionality (i.e., the number of nearby road segments that contribute to the prediction) to 100. From the perspective of spatial analysis, traffic congestion that occurs at a single location may propagate to other regions that are located a significant distance away from the congested site. This phenomenon has been witnessed and confirmed by [3, 4]. From the perspective of temporal analysis, a strong correlation usually exists among traffic time series, where previous traffic conditions likely have a large impact on future traffic. For a large-sized network with hundreds of links, capturing the spatial and temporal features of any link is very challenging [4, 5]. Due to the emergence of big data and deep learning, predicting large-scale network traffic has become feasible due to abundant traffic sensor data and hierarchical representations in deep architectures.

In the domain of computer vision, deep learning has achieved better performance than traditional image-processing paradigms. Deep learning in motion prediction is a research area in which the future movement of an object is predicted based on a series of historical scenes of the same object. Based on the success of this method, we snapshot network-wide traffic speeds as a collection of static images via a grid-based segmentation method, where each pixel represents the traffic condition of a single road segment or multiple road segments. As time evolves, the network-wide traffic prediction problem becomes a motion-prediction issue. Given a sequence of static images that comprise an animation, can we predict the future motion of each pixel? The deep-learning framework presents superior advantages in enhancing the motion prediction accuracy [6, 7]. Both spatial and temporal long-range dependencies should be considered when a video sequence is learned. Convolutional neural networks (CNNs) adopt layers with convolution filters to extract local features through sliding windows [8] and can model nearby or citywide spatial dependencies [9]. To learn time series with long time spans, long short-term memory (LSTM) neural networks (NNs), which were proposed by Hochreiter and Schmidhuber [10] in 1997, have been effectively applied in short-term traffic prediction [11, 12] and achieve remarkable performance in capturing the long-term temporal dependency of traffic flow. Motivated by the success of CNNs and LSTMs, this paper proposes a spatiotemporal image-based approach to predict the network-wide traffic state using spatiotemporal recurrent convolutional networks (SRCNs). Deep convolutional neural networks (DCNNs) are utilized to mine the space features among all links in an entire traffic network, whereas LSTMs are employed to learn the temporal features of traffic congestion evolution. We input the spatiotemporal features into a fully connected layer to learn the traffic speed pattern of each link in a large-scale traffic network and train the model from end to end.

The contributions of the paper can be summarized as follows:
- We developed a hybrid model named the SRCN that combines DCNNs and LSTMs to forecast network-wide traffic speeds.
- We proposed a novel traffic network representation method, which can retain the structure of the transport network at the fine scale.
- The special-temporal features of network traffic are modeled as a video, where each traffic condition is treated as one frame of the video. In the proposed SRCN architecture, the DCNNs capture the near- and far-side spatial dependencies from the perspective of the network, whereas the LSTMs learn the long-term temporal dependency. By the integration of DCNNs and LSTMs, we analyze spatiotemporal network-wide traffic data.

The remainder of this paper is organized as follows: Section 2 discusses the existing literature on traffic prediction. Section 3 introduces a grid-based transportation network representation approach for converting historical network traffic into a series of images and proposes the architecture of SRCNs to capture the spatiotemporal traffic features. In Section 4, a transportation network in Beijing with 278 links is employed to test the effectiveness of the proposed method. To evaluate the performance of SRCNs, we compare three prevailing deep learning architectures (i.e., LSTMs; DCNNs; and stacked auto encoders, SAEs) and a classical machine learning method (support vector machine, SVM). At the end of this paper, the conclusions are presented and future studies are discussed.

## 2. Literature review

Short-term traffic forecasting has attracted numerous researchers worldwide and can be traced to the 1970s. The approaches can be divided into two groups: parametric approaches and nonparametric approaches [13].

*2.1. Parametric approaches*

Parametric methods include the autoregressive integrated moving average (ARIMA), the Kalman filter (KF), and exponential smoothing (ES). Hamed et al. developed a simple ARIMA model of the order (0, 1, 1) to forecast the traffic volume on urban arterials [14]. Ding et al. classified the traffic modes into six classes and proposed a space-time autoregressive integrated moving average (STARIMA) model to forecast the traffic volume in urban areas in five-minute intervals [15]. S.R. Chandra and H. Al-Deek proposed vector autoregressive models for short-term traffic prediction on freeways that consider upstream and downstream location information and yield high accuracy [16]. Motivated by the superior capability to cast the regression problem of a KF, numerous KF-based traffic prediction studies began to emerge [17-19]. S.H. Hosseini et al. applied an adaptive neuro fuzzy inference system (ANFIS) based on KF to address the nonlinear problem of traffic speed forecasting [20]. B. Williams et al. developed a traffic flow prediction approach based on exponential smoothing, and K.Y. Chan employed a smoothing technique to pre-process traffic data before inputting the data into NNs for prediction, which achieved more than 6% accuracy [21, 22].

*2.2. Nonparametric approaches*

Compared with parametric approaches, nonparametric models are flexible and complex since their structure and parameters are not fixed. In the domain of nonparametric approaches, an SVM that is based on statistical learning theory is popular in the field of prediction [23]. The premise of an SVM is to map low-dimensional nonlinear data into a high-dimensional space by a kernel function. However, an SVM is highly sensitive to the choices of the kernel function and parameters. Many researchers have attempted to optimize an SVM and apply it to traffic prediction to derive some improved SVM variants, such as chaos wavelet analysis SVMs [24], least squares SVMs [25], particle swarm optimization SVMs [26], and genetic algorithm SVMs [27].

Another typical nonparametric method is an NN, which is extensively applied in almost every field, including traffic prediction. An NN can model complex nonlinear problems with remarkable performance in handling multi-dimensional data [28]. S.H. Huang et al. constructed an NN model to predict traffic speed that considers weather conditions [29]. A. Khotanzad and N. Sadek applied multilayer perceptron (MLP) and a fuzzy neural network (FNN) to high-speed network traffic prediction; the results indicate that NN performs better than the autoregressive model [30]. C. Qiu et al. developed a Bayesian-regularized NN to forecast short-term traffic speeds [31]. X. Ma [11] proposed a congestion prediction method that is based on RNN-RBM for a large-scale transportation network that included 515 road links.

In recent years, deep NNs, such as deep belief networks (DBNs), have been investigated in traffic flow prediction [32-35]. Although these methods are suitable for small-scale traffic networks or networks with few links, they fail to take advantage of correlations among different links and the long-term memory of traffic. To overcome these drawbacks, a special recurrent NN named LSTM is proposed to forecast traffic speed and traffic flow (X. Ma [36], Y. Tian [37], Y. Chen [38], R. Fu [39]); the results indicate that LSTMs outperform MLP and SVMs. The temporal features of traffic can be mined by time-series algorithms, such as LSTMs; however, these algorithms always fail to capture the spatial features among links. The capability of CNNs to extract spatial features in a local or city-wide region has been proven. Wu and Tan [40] constructed a short-term traffic flow prediction method based on the combination of CNNs and LSTMs on an arterial road. In this approach, the road is divided into serval links to view the road as a vector. This vector is input into one-dimensional CNNs to capture the spatial features of the links, and LSTMs are utilized to mine the temporal characteristics. This method can extract spatiotemporal correlations on a single arterial road but fails to consider ramps, interchanges, and intersections, which are significant components of any transportation network [41]. Consequently, the method disregards the spatial-propagation effect of congestion: a traffic incident that occurs on one link may influence the traffic conditions in far-side regions. Considering Figure. 1 as an example, the four-way intersection with four ramps has 25 links. If an incident occurs on link 9, then link 8, link 15 and link 21 are very likely to be congested. One-dimensional CNNs cannot adequately capture the spatial relations among links 8, 9, 15, and 21 because the convolutional filter of one-dimensional CNNs can only include a finite number of consecutive traffic speeds along each link and is unable to consider the zonal spatial dependencies among links that are not adjacent to each other, such as link 16 and link 3. In this circumstance, a two-dimensional (2D) convolutional filter must be employed to address regional traffic conditions. This improvement is especially important for predicting traffic at interchanges and intersections.

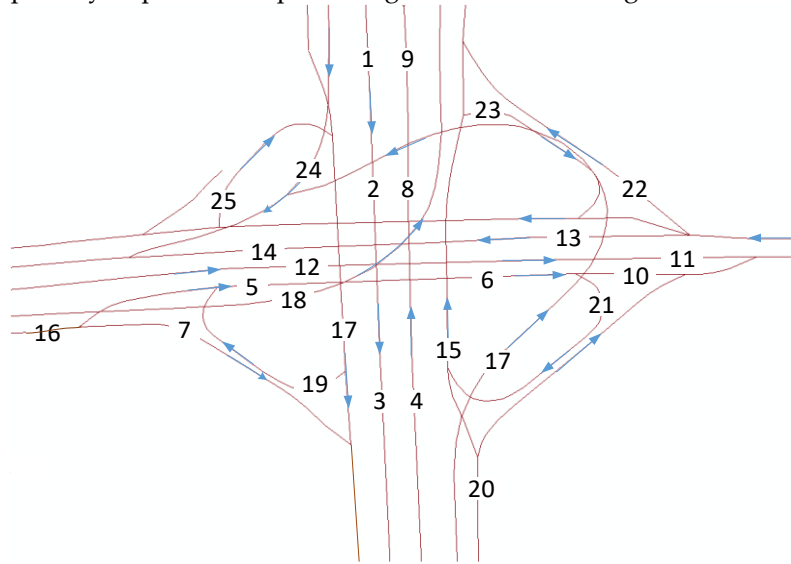

**Figure. 1.** Layout of an interchange in the Beijing Transportation Network

To address these drawbacks, based on the traffic network level, this paper proposes a novel NN structure that combines deep 2D CNNs and deep LSTMs to obtain the spatiotemporal correlations among all links in a traffic network. Several successful applications have verified the feasibility of combining CNNs and LSTMs, such as image description [42], visual activity recognition [43], and sentiment analysis [44]. Thus, we view the traffic network evolution process as a video, where every frame represents a traffic state and several future frames can be forecasted based on several previous frames. Based on this idea, the future traffic state can be effectively forecasted using well-established image-processing algorithms.

## 3. Methodology

In this section, we construct our SRCNs for predicting the traffic state. Figure. 7 provides a graphical illustration of the proposed model. An SRCN consists of a 2D CNN and two LSTMs; the details are presented in the following section.

### 3.1. Network representation

Assume that we want to predict the congestion at every link in a traffic network. We establish the links $\{y_i\}_{i=1}^{n}$, where *n* represents the total number of links.

Step 1: We choose a traffic network (refer to Figure. 2A), divide it into *n* links, and calculate the average speeds on these links over a particular time period, which is set to two minutes according to Eq. 1, where *m* and $\bar{v}_j$ represent the number of vehicles and their average speed, respectively, on link j. We map the calculated speeds on the links to different colors (as shown in Figure. 2B). Figure. 3A shows an example of a small-scale network.

$$v_i = \frac{\sum_{j=1}^{m}\bar{v}_j}{m} \quad (1)$$

Step 2: Divide the traffic network using a small grid, whose size is (0.0001°×0.0001°, longitude and latitude), where 0.0001 in longitude (or latitude) in Beijing is equal to approximately 10 meters (shown in Figure. 3B), and each grid box represents a spatial region.

Step 3: Map the average speed to the grid. The value of a blank area is set to zero; if multiple links pass through the same grid box, we assign their average speed to the box (as shown in Figure. 3C) and scale the speed to (0, 1) (as shown in Figure. 2C).

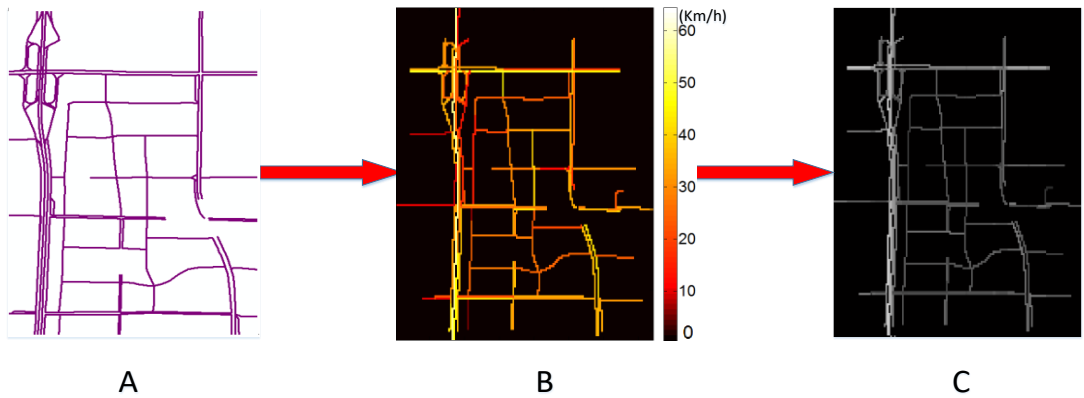

**Figure. 2.** Grid-based transportation network segmentation process

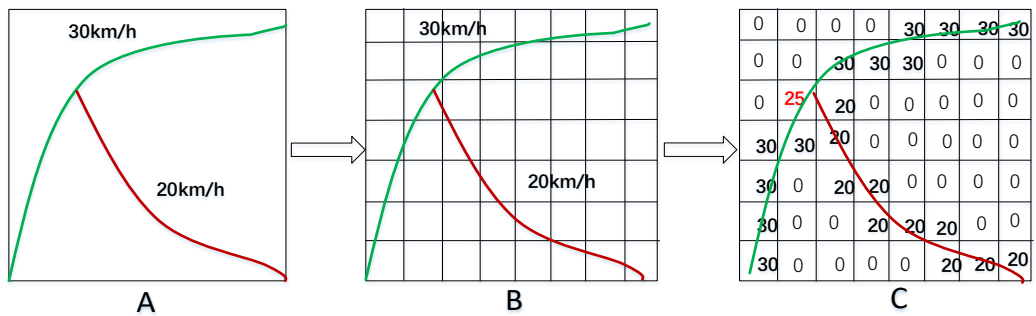

**Figure. 3.** Traffic speed representation in a small-scale transportation network

Using the grid-based network-segmentation method, the relative topology among different links remains unchanged. This treatment can retain the geometric information of roads, such as sharp U-turns and interchanges at fine granularity.

*3.2. Spatial features captured by a CNN*

The congestion in one link not only affects its most adjacent links but also may propagate to other far-side regions. CNNs have been successful in extracting features. In this study, we construct deep convolutional neural networks (DCNNs) to capture the spatial relationships among links. The spatial dependencies of nearby links (two blue lines or red lines in Figure. 4A) can be mined by the shallow convolutional layer and the spatial dependencies for more distant links (red and blue lines in Figure. 4A) can be extracted by the deep convolutional layer, because the distance among them will be shortened due to the convolution and pooling processes (shown in Figure. 4 A to D). For example, the purple lines in Figure. 4 A represent a small-scale traffic network, and the blue and red grid boxes in Figure. 4 B indicate the near-spatial dependency captured by the shallow convolutional layer. We discover that the distance between the two regions in Figure. 4(A) decreases as additional convolutional and pooling operations become involved. The gray grid in Figure. 4D represents the far-spatial dependency mined by the deep convolutional layers. These abstract features are significant for the prediction problem [45].

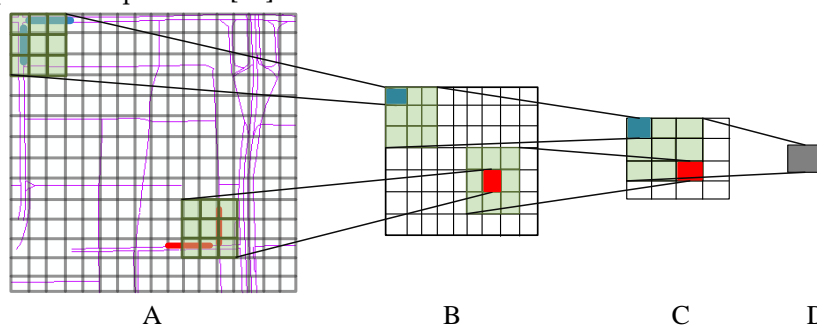

**Figure. 4.** Convolutions for capturing near and far dependencies. Each grid box represents a spatial region (similar to the network representation in step 2), the transparent green region represents a $3\times 3$ convolutional filter, the two blue lines in A represent two nearby links, and the blue line and red line in A represent two additional distant links.

We naturally utilize a 2D CNN to capture the spatial features of the traffic network. The input for DCNNs is an image (Figure. 2C) that represents one traffic state, and the pixel values in the image range from 0 to 1. The network framework is shown as Figure. 5, including the input layer, convolution layer, pooling layer, fully connected layer and output layer. The details of each part are subsequently explained.

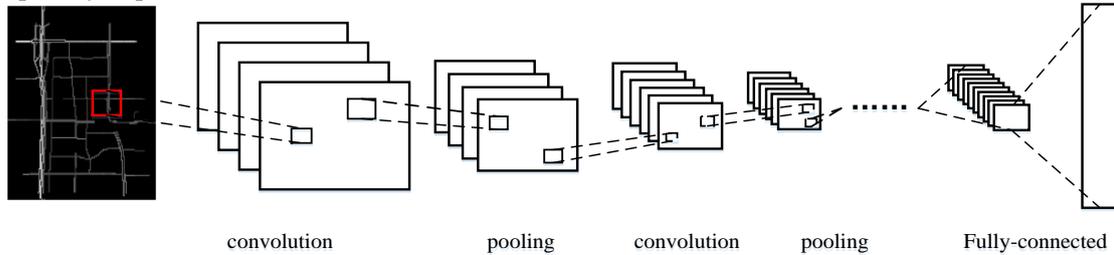

convolution     pooling     convolution     pooling     Fully-connected

**Figure. 5.** Structure of the DCNNs

The input image at time $t$ to the DCNNs is set to $A^T = \{a_{m,n}^t\}$, where m and n represent the latitude coordinates and longitude coordinates, respectively, and the output of the DCNNs at time $t$ is set to $X^t = \{x_u^t\}_{u=1}^{p}$, where $p$ is the number of links in the traffic network. The feature extraction is performed by convolving the input with filters. Denote the $r-th$ filter output of the $l-th$ layer as $O_r^l$, and denote the $k-th$ filter output of the previous layer as $O_r^{l-1}$. Thus, $O_r^l$ can be calculated by Eq. 2, where $W_{kr}^l$ and $b_r^l$ are the weight and the bias, $*$ denotes the convolution operation, and $f$ is a nonlinear activation function. After convolution, max-pooling is employed to select the salient features from the receptive region and to greatly reduce the number of model parameters by merging groups of neurons.

$$O_r^l = f\left(\sum_k W_{kr}^l * O_r^{l-1} + b_r^l\right) \quad (2)$$

*3.3. Long short-term temporal features*

Traffic data has a distinct temporal dependency, such as video and language, and the traffic state several hours earlier may have a long-term impact on the current state. The most successful model for handling long-term time series prediction is LSTM, which achieves powerful learning ability by enforcing constant error flow through designed special units [46]. Traditional RNNs suffer from vanishing or exploding gradients when the number of time steps is large. LSTMs introduce memory units to learn whether to forget previous hidden states and update hidden states; they have been shown to be more effective than traditional RNNs [47]. Motivated by the temporal dynamics of traffic flow and the superior performance in long-term time-series prediction, we explore the application of LSTMs as a key component in predicting spatiotemporal traffic speeds in a large-scale transportation network.

LSTMs are considered to be a specific form of RNNs; each LSTM is composed of one input layer, one or several hidden layers and one output layer. The key to LSTMs is a memory cell, which is employed to overcome the vanishing and exploding gradients in traditional RNNs. As shown in Figure. 6, the LSTMs contains three gates, namely, the input gate, forget gate and output gate. These gates are used to decide whether to remove or add information to a cell state.

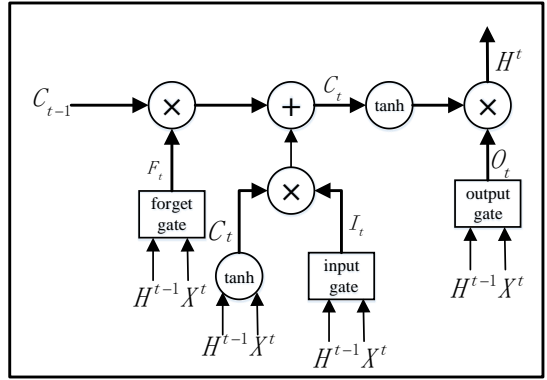

**Figure. 6.** LSTM NN architecture

In this paper, at time $t$, the output $X^t = \{x_u^t\}_{u=1}^{p}$ of DCNNs represents the input of LSTMs, and the output of LSTMs is denoted as $H_t = \{h_u^t\}_{u=1}^{q}$, where $q$ represents the number of hidden units. The cell input state is $\tilde{C}_t$, the cell output state is $C_t$, and the three gates' states are $I_t, F_t, O_t$. The temporal features of the traffic state will be iteratively calculated according to Eqs. 3-8:

Input gate: $\quad I_t = \sigma\left(W_1^i X_q + W_h^i H_{t-1} + b_i\right),$  (3)

Forget gate: $\quad F_t = \sigma\left(W_1^f X_t + W_h^f H_{t-1} + b_f\right),$  (4)

Output gate: $\quad O_t = \sigma\left(W_1^o X_t + W_h^o H_{t-1} + b_o\right),$  (5)

Cell input: $\quad \tilde{C}_t = \tanh\left(W_1^c X_t + W_h^c H_{t-1} + b_c\right),$  (6)

Cell output: $\quad C_t = I_t \odot \tilde{C}_t + F_t \odot C_{t-1},$  (7)

Hidden layer output: $\quad H_t = O_t \odot \tanh(C_t),$  (8)

where $W_1^i, W_1^f, W_1^o, W_1^c$ are the weight matrices that connect $X^t$ to the three gates and the cell input, $W_h^i, W_h^f, W_h^o, W_h^c$ are the weight matrices that connect $H^{t-1}$ to the three gates and the cell input, $b_i, b_f, b_o, b_c$ are the biases of the three gates and the cell input, $\sigma$ represents the sigmoid function, tanh represents the hyperbolic tangent function, and $\odot$ represents the scalar product of two vectors.

*3.4. Spatiotemporal Recurrent Convolutional Networks*

The spatiotemporal features of the traffic state can be learned by CNNs and LSTMs. The next step is to forecast the future traffic state by the integration of CNNs and LSTMs. The output of LSTMs is utilized as an input to a fully connected layer. The predicted speed value is calculated by Eq. 9, where $W_2, b$ represent the weight and the bias between the hidden layer and the fully connected layer, respectively, which are the output of the entire model; and we train the model from end to end.

$$Y^{t+1} = W_2 \times H_t + b_2 \quad (9)$$

In this section, we propose a novel deep architecture named a spatiotemporal recurrent convolutional network (SRCN) to predict the network-wide traffic state. A graphical illustration of the proposed model is shown in Figure. 7. Each SRCN consists of a DCNN, two LSTMs, and a fully connected layer. The detailed structure of the SRCNs is described in the experiment section. The values of $a, b, c$ in Figure. 4 can be arbitrarily established, which indicates that we can make multi-steps predictions; for example, if we set $a, b, c$ to (2, 4, 5), we can predict the traffic states of the next (2, 4, 5) time steps based on the historical data of several steps.

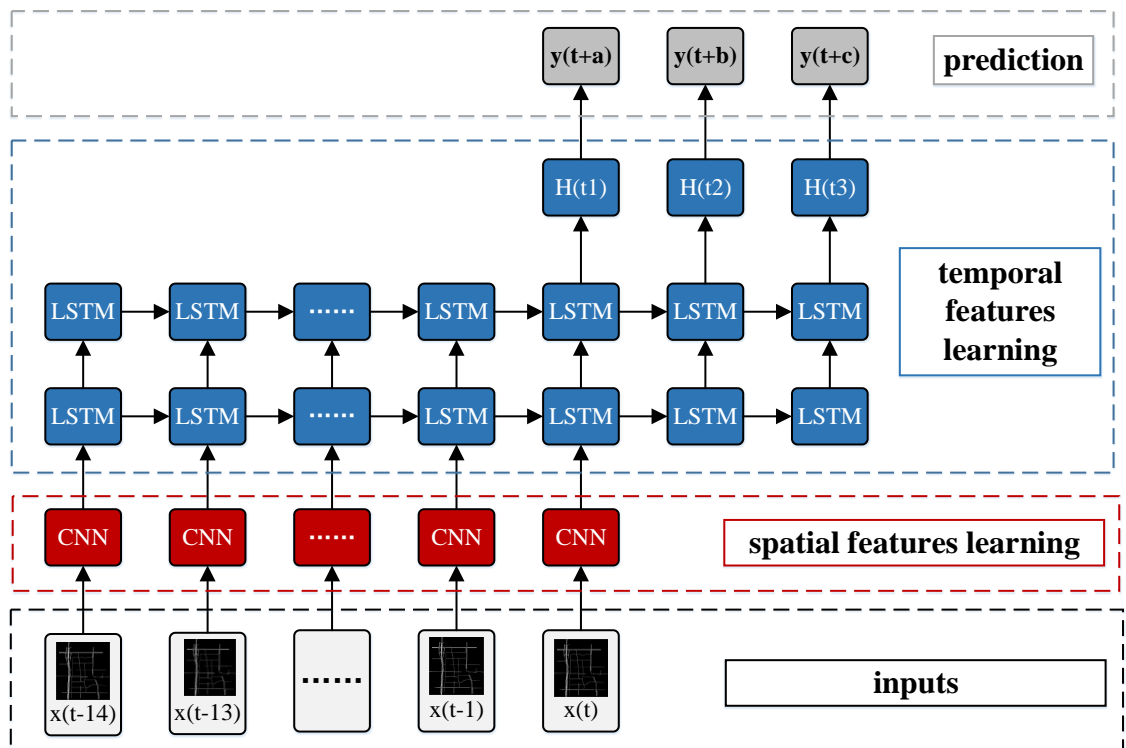

**Figure. 7.** Framework of SRCNs

## 4. Methodology

*4.1. Data Source*

Data were collected from Jun. 1, 2015, to Aug. 31, 2015, (92 days) with an updating frequency of two minutes; the time period in this study ranges from 6:00:00 to 22:00:00, when high travel demand is commonly observed. Thus, 481 traffic states exist per day. The traffic network is located between the Second Ring Road and Third Ring Road in Beijing. The network encompasses 278 links, and the total length of the network exceeds 38.486 km, including seven arterial roads and hundreds of interchanges and intersections. The data were divided into two subsets: data from the first two months were employed for training, and the remaining data were employed for testing.

For all methods, the time lag is set to 15, which indicates that the traffic states of the previous min were used to predict the traffic states at time steps $15 \times 2 = 30$. For example, if $t+a, t+b, t+c$, which indicates that historical data from the previous 30 min are used to predict the traffic state 30 min in the future. Different settings are tested in the following experiments.

*4.1. Implementation*

The details of our SRCNs are shown in Table 1. SRCNs are trained based on the optimizer RMSprop, which has been proven to work well [48], especially in the RNN model [49]. The learning rate is set to 0.003; the decay parameter is set to 0.9; and the batch size is set to 64. The loss function is the mean squared error (MSE) and the validation-data proportion is set to 20%. A batch-normalization layer is used to overcome internal covariate shift. Since our model is very deep, we can also employ a substantially higher learning rate to accelerate convergence [50]. The dropout layer and early stopping are used to prevent overfitting [51], and all parameters of our model are dependent on numerous experiments to yield an optimal structure. The structures of other methods (LSTMs and SAEs) are established according to their papers.

**Table 1.** Parameter settings of SRCNs

| Layer | Name | Channels | size |
| --- | --- | --- | --- |
| **0** | Inputs | 1 | (163,148) |
| **1** | Convolution | 16 | (3,3) |
| | Max-pooling | 16 | (2,2) |
| | Activation (relu) | —— | —— |
| | Batch-normalization | —— | —— |
| **2** | Convolution | 32 | (3,3) |
| | Max-pooling | 32 | (2,2) |
| | Activation (relu) | —— | —— |
| | Batch-normalization | —— | —— |
| **3** | Convolution | 64 | (3,3) |
| | Activation (relu) | —— | —— |
| | Batch-normalization | —— | —— |
| **4** | Convolution | 64 | (3,3) |
| | Activation (relu) | —— | —— |
| | Batch-normalization | —— | —— |
| **5** | Convolution | 128 | (3,3) |
| | Max-pooling | 128 | (2,2) |
| | Activation (relu) | —— | —— |
| | Batch-normalization | —— | —— |
| **6** | Flatten | —— | —— |
| **7** | Fully connected | —— | 278 |
| **8** | Lstm1 | —— | 800 |
| | Activation (tanh) | —— | |
| **9** | Lstm2 | —— | 800 |
| | Activation (tanh) | —— | —— |
| **10** | Dropout (0.2) | —— | —— |
| **11** | Fully connected | —— | 278 |

*4.1. Comparison and Analysis of Results*

In this section, we employ traffic speed data from Beijing, China to evaluate our model—SRCNs—and compare them with other deep NNs, including LSTMs [36], SAEs [52], DCNNs, and SVM. For the DCNN model, the structure is the same as the first part of the SRCNs. For the SVM model, the kernel function is the radial basis function (RBF), and the trade-off parameter "c" and width parameter "g" are calibrated using five-fold cross validation. For comparison and analysis, we specify two different conditions: $(a,b,c)=(1,2,3)$ for short-term prediction and $(a,b,c)=(10,20,30)$ for long-term prediction. The mean absolute percentage error (MAPE) and root mean squared error (MSE) are utilized to measure the performance of traffic state forecasting in this paper, which are defined in Eqs. 10-11, where $y_{it}$ and $z_{it}$ denote the predicted traffic speeds and actual traffic speeds, respectively, at time $t$ at location $i$, $m$ is the total number of predictions, and $n_p = m*n$. In the experiment, the value of $n$ is 278, and the value of $m$ is 14896, which indicates that we test 278 links and 14896 traffic states.

$$MAPE = \frac{1}{n_p}\sum_{i=1}^{n}\sum_{t=1}^{m}\left(\frac{y_{it}-z_{it}}{y_{it}}\right) \tag{10}$$

$$RMSE = \sqrt{\frac{1}{n_p} \sum_{i=1}^{n} \sum_{t=1}^{m} (y_{it} - z_{it})^2} \qquad (11)$$

*4.2. Short-term prediction*

Short-term prediction is primarily employed for en-route trip planning and is desired by travelers who resort to in-vehicle navigation devices. In this section, we set $(a,b,c) = (1,2,3)$, which indicates that we will predict traffic speeds in the next (2, 4, 6) min based on historical data from the previous 30 min. The results of SRCNs, LSTMs, SAEs, DCNNs, and SVM are listed in Figure. 8 and Table 2.

**Table 2.** Comparison among different methods in terms of short-term prediction

| Time steps | 2 min | | 4 min | | 6 min | | Average error | |
|---|---|---|---|---|---|---|---|---|
| Algorithm | MAPE | RMSE | MAPE | RMSE | MAPE | RMSE | MAPE | RMSE |
| **SRCNs** | **0.1269** | **4.9258** | **0.1271** | **5.0124** | **0.1272** | **5.0612** | **0.1270** | **4.9998** |
| LSTMs | 0.1630 | 6.1521 | 0.1731 | 6.8721 | 0.1781 | 7.0016 | 0.1714 | 6.7527 |
| SAEs | 0.1591 | 6.2319 | 0.1718 | 6.8737 | 0.1742 | 7.2602 | 0.1684 | 6.7886 |
| DCNNs | 0.1622 | 6.6509 | 0.1724 | 6.8516 | 0.1775 | 7.2845 | 0.1707 | 6.9290 |
| SVM | 0.1803 | 7.6036 | 0.2016 | 8.0132 | 0.2123 | 8.2346 | 0.1984 | 7.9505 |

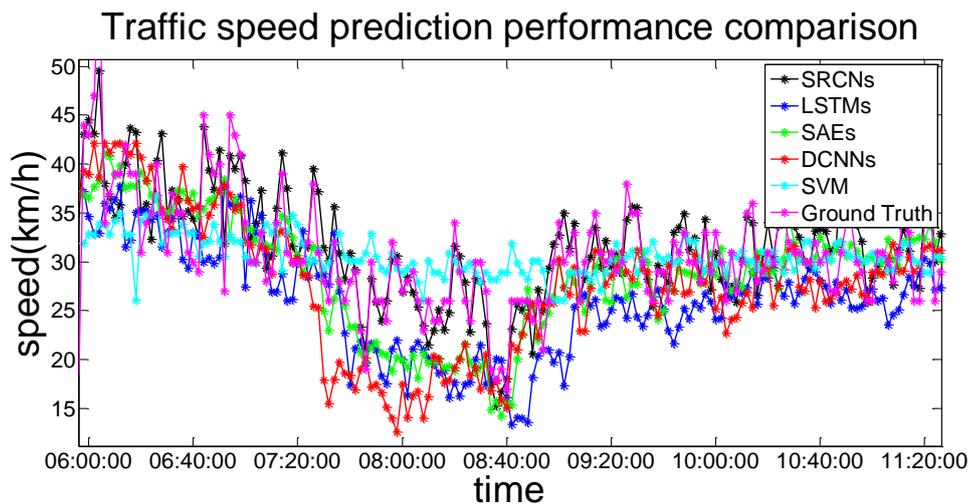

**Figure. 8.** Traffic speed prediction performance comparison at 2 min time steps

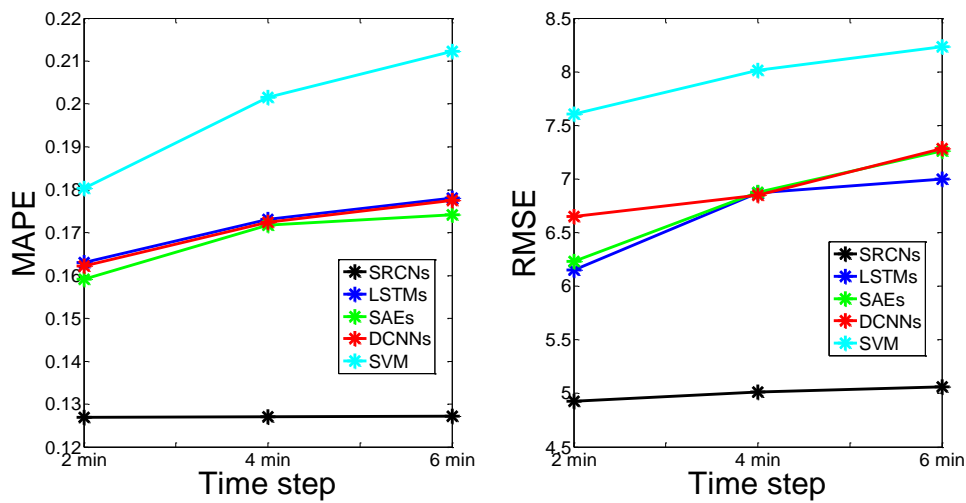

**Figure. 9.** Prediction errors for prediction horizons of (2, 4, and 6) min in terms of the MAPE and RMSE

In this section, we compare SRCNs with other four algorithms (LSTMs, SAEs, DCNNs, and SVM). We observe that SRCNs yield the most accurate results for short-term traffic speed prediction in terms of MAPE and RMSE; the results are shown as Figure. 9. One possible reason is that SRCNs consider spatiotemporal features. In the order listed in Table 2, the average MAPE values for the other algorithms decrease by 34.96%, 32.60%, 34.41% and 56.22%, and the average RMSE values for the other algorithms decrease by 35.06%, 35.78%, 38.59%, and 59.02%. The SVM model exhibits the worst prediction performance, and the LSTMs and DCNNs show similar precision, which indicates that spatial and temporal features have similar roles in short-term traffic prediction. The MAPE of SRCNs is approximately 0.1, and the RMSE is approximately 5. As shown in Figure. 9, we determine that the prediction error increases as the prediction horizon increases. SRCNs yield the lowest prediction error with a stable trend.

*4.2. Long-term prediction*

Long-term prediction, which is primarily adopted by pre-route travelers who plan their trips in advance, is considered to be more challenging than short-term prediction. In this section, we set , which indicates that we will predict the traffic speed in the next (20, 40, 60) min based on historical data for the previous 30 min. The results of SRCNs, LSTMs, SAEs, DCNNs, and SVM are listed in Figure. 10 and Table 3.

**Table 3.** Comparison among different methods in terms of long-term prediction

| Time steps<br>Algorithm | 20 min | | 40 min | | 60 min | | Average error | |
|---|---|---|---|---|---|---|---|---|
| | MAPE | RMSE | MAPE | RMSE | MAPE | RMSE | MAPE | RMSE |
| **SRCNs** | **0.1661** | **6.0569** | **0.1753** | **6.5631** | **0.1889** | **6.8516** | **0.1768** | **6.4905** |
| **LSTMs** | 0.1700 | 7.1857 | 0.1872 | 7.7322 | 0.2003 | 7.9843 | 0.1858 | 7.6340 |
| **SAEs** | 0.2045 | 7.2374 | 0.2139 | 7.9737 | 0.2228 | 8.2881 | 0.2137 | 7.8331 |
| **DCNNs** | 0.2018 | 7.6608 | 0.2531 | 8.8613 | 0.3264 | 12.5846 | 0.2604 | 9.7022 |
| **SVM** | 0.3469 | 12.9577 | 0.3480 | 13.1810 | 0.3621 | 13.4676 | 0.3542 | 13.2021 |

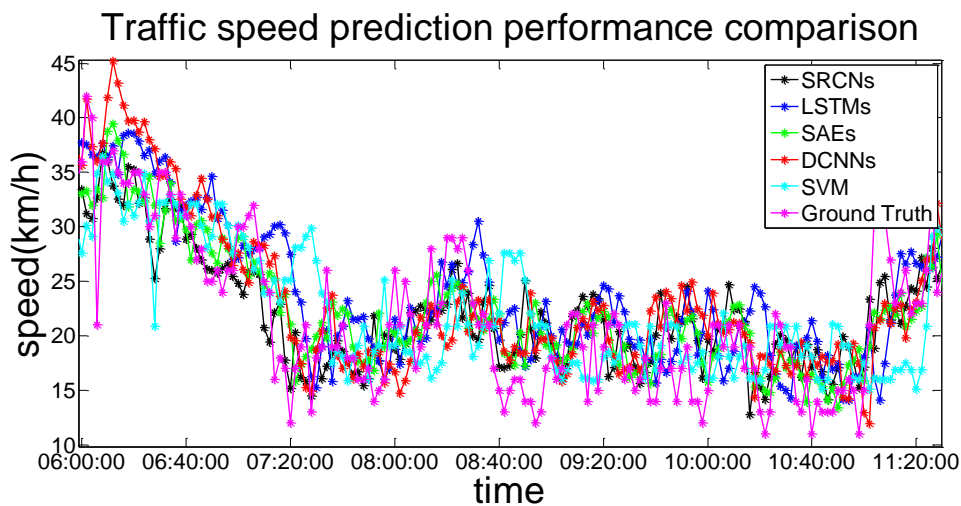

**Figure. 10.** Traffic speed prediction performance comparison at 20 min prediction horizon

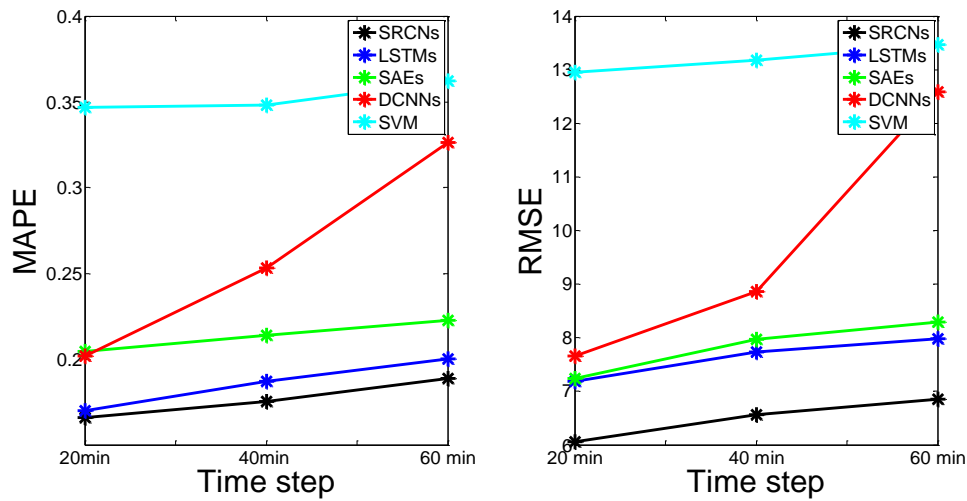

**Figure. 11.** Prediction errors for prediction horizons (20, 40, and 60) min in terms of MAPE and RMSE

In this section, we compare SRCNs with four algorithms (LSTMs, SAEs, DCNNs, and SVM), as detailed in Table 2. We determine that the SRCN model is superior to other models in long-term traffic speed prediction in terms of MAPE and RMSE, as shown in Figure. 11. This finding confirms the advantages of SRCNs in utilizing the spatiotemporal features in traffic networks. The average MAPE values for the other algorithms, in the order listed in Table 3, decrease by nearly 18.58%, 20.87%, 47.29% and 100.34% relative to the SRCNs in long-term prediction. The average RMSE values for the other algorithms decrease by 17.62%, 20.69%, 49.45%, and 103.41%. Similar to short-term prediction, the SVM model exhibits the worst prediction performance. However, LSTMs perform much better than DCNNs, which indicates that spatial information contributes more than temporal features for long-term traffic prediction. The SRCNs outperform other algorithms, with the lowest MAPE—approximately 0.2—and an RMSE of approximately 6. As shown in Figure. 11, we discover that the error increases as the prediction horizon increases, but the long-term prediction performance decays more rapidly than the short-term prediction performance.

SRCNs achieve the best accuracy compared with the other four algorithms (LSTMs, SAEs, DCNNs, and SVM) in both short- and long-term traffic speed prediction and obtain the most stable error trend, because SRCNs can learn both spatial and temporal features on a network-wide scale. SRCNs can perform multi-step-ahead prediction due to the special structure of LSTMs and can output a sequence of predictions [53]. These results verify the superiority and feasibility of the SRCNs, which employ deep CNNs to capture the special features and mine temporal regularity using a LSTM NN.

**5. Conclusions and Future Studies**

Inspired by the research findings of motion prediction in the domain of computer vision, where the future movement of an object can be estimated from a sequence of scenes generated by the same object, we proposed a novel grid-based transportation network segmentation method. The network-wide traffic can be snapshot as a series of static images and can retain the complicated road network topology, including interchanges, intersections and ramps. Based on the proposed network representation method, a novel deep learning architecture named SRCN is developed; this method inherits the advantages of both DCNNs and LSTMs. DCNNs are employed to capture the near- and far-side spatial dependencies among different links, and LSTMs are utilized to learn the long-term temporal dependency of each link. To validate the effectiveness of the proposed SRCNs, traffic speed data were collected for three months with an updating frequency of 2 min from a Beijing transportation network with 278 links. Data from the first two months were employed for training, and the remaining data were employed for testing. In addition, three prevailing deep learning NNs (i.e., LSTMs, DCNNs, and SAEs) and a classical machine learning method (SVM) were compared with the SRCNs for the same dataset. The numerical experiments demonstrate that the SRCNs outperformed other algorithms in terms of accuracy and stability, which indicates the potential of combining DCNNs with LSTMs for large-scale network-wide traffic prediction applications.

In future studies, the model can be improved by considering additional factors, such as weather, social events, and traffic control. The training efficiency can be enhanced by optimizing pre-training methods, which may reduce the number of iterations while achieving more accurate results. Another intriguing research direction is to develop novel transportation network representation approaches. By eliminating the blank regions without any roadway network, the computational burden of training SRCNs should be greatly reduced. In addition, we aim to expand the transportation network to a larger scale.

**Acknowledgments:** This paper is supported by the National Natural Science Foundation of China (51308021, 51408019 and U1564212), Beijing Nova Program (z151100000315048), Beijing Natural Science Foundation (9172011) and Young Elite Scientist Sponsorship Program by the China Association for Science and Technology.

**Author Contributions:** Xiaolei Ma contributed analysis tools and the idea; Zhihai Wu and Haiyang Yu performed the experiments and wrote the paper; Shuqin Wang was in charge of the final version of the paper; Yunpeng Wang collected and processed the data.

**Conflicts of Interest:** The authors declare no conflict of interest.

**References**

1. M.T. Asif, J. Dauwels, C.Y. Goh, and A. Oran, Spatiotemporal Patterns in Large-Scale Traffic Speed Prediction, *IEEE Transactions on Intelligent Transportation Systems* **2014**, pp. 794-804.
2. X. Ma, C. Liu, H. Wen, Y. Wang, and Y.J. Wu, Understanding commuting patterns using transit smart card data, *JOURNAL OF TRANSPORT GEOGRAPHY* 2017, pp. 135-145.
3. S. Yang, On feature selection for traffic congestion prediction, *Transportation Research Part C* **2013**, pp. 160-169.
4. J. Zhang, Y. Zheng, D. Qi, R. Li, and X. Yi, DNN-based prediction model for spatio-temporal data, *Proc. ACM Sigspatial International Conference on Advances in Geographic Information Systems*, 2016, pp. 92.
5. Z. He, L. Zheng, P. Chen, and W. Guan, Mapping to Cells: A Simple Method to Extract Traffic Dynamics from Probe Vehicle Data, *Computer‐aided Civil & Infrastructure Engineering* **2017**.
6. J. Walker, A. Gupta, and M. Hebert, Dense Optical Flow Prediction from a Static Image, 2015.
7. M.S. Pavel, H. Schulz, and S. Behnke, Object class segmentation of RGB-D video using recurrent convolutional neural networks., *Neural Networks* **2017**, pp. 105-113.
8. P. Qin, W. Xu, and J. Guo, An empirical convolutional neural network approach for semantic relation classification, *Neurocomputing* **2016**, pp. 1-9.
9. J. Zhang, Y. Zheng, and D. Qi, Deep Spatio-Temporal Residual Networks for Citywide Crowd Flows Prediction, 2016.
10. S. Hochreiter, and J. Schmidhuber, Long short-term memory, *Neural Computation* **1997**, pp. 1735-1780.
11. X. Ma, H. Yu, Y. Wang, and Y. Wang, Large-Scale Transportation Network Congestion Evolution Prediction Using Deep Learning Theory, *PLoS One* **2015**, pp. e119044.
12. Z. Zhao, W. Chen, X. Wu, P.C. Chen, and J. Liu, LSTM network: a deep learning approach for short-term traffic forecast, *IET Intelligent Transport Systems* **2017**, pp. 68-75.
13. H. Van Lint, and C. Van Hinsbergen, Short-Term Traffic and Travel Time Prediction Models, *Transportation Research E-Circular* **2012**.
14. M.M. Hamed, H.R. Al-Masaeid, and Z.M.B. Said, Short-Term Prediction of Traffic Volume in Urban Arterials, *Journal of Transportation Engineering* **1995**, pp. 249-254.
15. Q.Y. Ding, X.F. Wang, X.Y. Zhang, and Z.Q. Sun, Forecasting Traffic Volume with Space-Time ARIMA Model, *Advanced Materials Research* **2010**, pp. 979-983.
16. S.R. Chandra, and H. Al-Deek, Predictions of Freeway Traffic Speeds and Volumes Using Vector Autoregressive Models, *Journal of Intelligent Transportation Systems* **2009**, pp. 53-72.
17. J. Guo, W. Huang, and B.M. Williams, Adaptive Kalman filter approach for stochastic short-term traffic flow rate prediction and uncertainty quantification, *Transportation Research Part C Emerging Technologies* **2014**, pp. 50-64.
18. J.W.C. Van Lint, Online Learning Solutions for Freeway Travel Time Prediction, *Intelligent Transportation Systems IEEE Transactions on* **2008**, pp. 38-47.
19. H. Chen, and S. Grant-Muller, Use of sequential learning for short-term traffic flow forecasting, *Transportation Research Part C Emerging Technologies* **2001**, pp. 319-336.
20. S.H. Hosseini, B. Moshiri, A. Rahimi-Kian, and B.N. Araabi, Short-term traffic flow forecasting by mutual information and artificial neural networks, *Proc. IEEE International Conference on Industrial Technology*, 2012, pp. 1136-1141.

21. B. Williams, P. Durvasula, and D. Brown, Urban Freeway Traffic Flow Prediction: Application of Seasonal Autoregressive Integrated Moving Average and Exponential Smoothing Models, *Transportation Research Record* **1998**, pp. 132-141.
22. K.Y. Chan, T.S. Dillon, J. Singh, and E. Chang, Neural-Network-Based Models for Short-Term Traffic Flow Forecasting Using a Hybrid Exponential Smoothing and Levenberg–Marquardt Algorithm, *IEEE Transactions on Intelligent Transportation Systems* **2012**, pp. 644-654.
23. T. Evgeniou, M. Pontil, and T. Poggio, Regularization Networks and Support Vector Machines, *Advances in Computational Mathematics* **2000**, pp. 1-50.
24. J. Wang, and Q. Shi, Short-term traffic speed forecasting hybrid model based on Chaos–Wavelet Analysis-Support Vector Machine theory, *Transportation Research Part C: Emerging Technologies* **2013**, pp. 219-232.
25. Y. Cong, J. Wang, and X. Li, Traffic Flow Forecasting by a Least Squares Support Vector Machine with a Fruit Fly Optimization Algorithm, *Procedia Engineering* **2016**, pp. 59-68.
26. Y. Gu, D. Wei, and M. Zhao, A New Intelligent Model for Short Time Traffic Flow Prediction via EMD and PSO–SVM, *Lecture Notes in Electrical Engineering* **2012**, pp. 59-66.
27. Z. Yang, D. Mei, Q. Yang, H. Zhou, and X. Li, Traffic Flow Prediction Model for Large-Scale Road Network Based on Cloud Computing, *Mathematical Problems in Engineering* **2014**, pp. 1-8.
28. M.G. Karlaftis, and E.I. Vlahogianni, Statistical methods versus neural networks in transportation research: Differences, similarities and some insights, *Transportation Research Part C Emerging Technologies* **2011**, pp. 387-399.
29. S.H. Huang, and B. Ran, An Application of Neural Network on Traffic Speed Prediction Under Adverse Weather Condition, *Transportation Research Board Annual Meeting* **2003**.
30. A. Khotanzad, and N. Sadek, Multi-scale high-speed network traffic prediction using combination of neural networks, *Proc. International Joint Conference on Neural Networks*, 2003, pp. 1071-1075.
31. C. Qiu, C. Wang, X. Zuo, and B. Fang, A Bayesian regularized neural network approach to short-term traffic speed prediction, 2011, pp. 2215-2220.
32. W. Huang, G. Song, H. Hong, and K. Xie, Deep Architecture for Traffic Flow Prediction: Deep Belief Networks With Multitask Learning, *IEEE Transactions on Intelligent Transportation Systems* **2014**, pp. 2191-2201.
33. Y. Lv, Y. Duan, W. Kang, and Z. Li, Traffic Flow Prediction With Big Data: A Deep Learning Approach, *IEEE Transactions on Intelligent Transportation Systems* **2015**, pp. 865-873.
34. H. Tan, X. Xuan, Y. Wu, Z. Zhong, and B. Ran, A Comparison of Traffic Flow Prediction Methods Based on DBN, *Proc. Cota International Conference of Transportation Professionals*, 2016, pp. 273-283.
35. H.F. Yang, T.S. Dillon, and Y.P. Chen, Optimized Structure of the Traffic Flow Forecasting Model With a Deep Learning Approach, *IEEE Transactions on Neural Networks & Learning Systems* **2016**, pp. 1-11.
36. X. Ma, Z. Tao, Y. Wang, H. Yu, and Y. Wang, Long short-term memory neural network for traffic speed prediction using remote microwave sensor data, *Transportation Research Part C Emerging Technologies* **2015**, pp. 187-197.
37. Y. Tian, and L. Pan, Predicting Short-Term Traffic Flow by Long Short-Term Memory Recurrent Neural Network, *Proc. IEEE International Conference on Smart City socialcom sustaincom*, 2015, pp. 153-158.

38. Y. Chen, Y. Lv, Z. Li, and F. Wang, Long short-term memory model for traffic congestion prediction with online open data, *Proc. Intelligent Transportation Systems (ITSC), 2016 IEEE 19th International Conference on*, IEEE, 2016, pp. 132-137.
39. R. Fu, Z. Zhang, and L. Li, Using LSTM and GRU neural network methods for traffic flow prediction, *Proc. Chinese Association of Automation (YAC), Youth Academic Annual Conference of*, IEEE, 2016, pp. 324-328.
40. Y. Wu, and H. Tan, Short-term traffic flow forecasting with spatial-temporal correlation in a hybrid deep learning framework, 2016.
41. Z. He, and L. Zheng, Visualizing Traffic Dynamics Based on Floating Car Data, *Journal of Transportation Engineering, Part A: Systems* **2017**, pp. 4017005.
42. O. Vinyals, A. Toshev, S. Bengio, and D. Erhan, Show and tell: A neural image caption generator, 2014, pp. 3156-3164.
43. J. Donahue, L.A. Hendricks, S. Guadarrama, M. Rohrbach, S. Venugopalan, T. Darrell, and K. Saenko, *Long-term recurrent convolutional networks for visual recognition and description*, Elsevier, 2014.
44. J. Wang, L.C. Yu, K.R. Lai, and X. Zhang, Dimensional Sentiment Analysis Using a Regional CNN-LSTM Model, *Proc. Meeting of the Association for Computational Linguistics*, 2016, pp. 225-230.
45. Y. Lecun, and Y. Bengio, Convolutional Networks for Images, Speech, and Time-Series, *Proc. The Handbook of Brain Theory and Neural Networks*, 1995.
46. S. Hochreiter, and J. Schmidhuber, Long Short-Term Memory, *Neural Computation* **1997**, pp. 1735-1780.
47. Y. Lecun, Y. Bengio, and G. Hinton, Deep learning, *Nature* **2015**, pp. 436-444.
48. T. Schaul, I. Antonoglou, and D. Silver, Unit Tests for Stochastic Optimization, *Nihon Naika Gakkai Zasshi the Journal of the Japanese Society of Internal Medicine* **2014**, pp. 1474-1483.
49. B.L. Sturm, J.F. Santos, O. Bental, and I. Korshunova, Music transcription modelling and composition using deep learning, 2016.
50. S. Ioffe, and C. Szegedy, Batch Normalization: Accelerating Deep Network Training by Reducing Internal Covariate Shift, *Computer Science* **2015**.
51. X. Ma, Z. Dai, Z. He, J. Ma, Y. Wang, and Y. Wang, Learning traffic as images: a deep convolutional neural network for large-scale transportation network speed prediction, *Sensors* **2017**, pp. 818.
52. Y. Lv, Y. Duan, W. Kang, and Z. Li, Traffic Flow Prediction With Big Data: A Deep Learning Approach, *IEEE Transactions on Intelligent Transportation Systems* **2015**, pp. 865-873.
53. J. Ryan, A.J. Summerville, M. Mateas, and N. Wardrip-Fruin, *Translating Player Dialogue into Meaning Representations Using LSTMs*, Springer International Publishing, 2016.